# AI applications in forest monitoring need remote sensing benchmark datasets

Emily R. Lines
*Department of Geography*
*University of Cambridge*
Cambridge, UK
erl27@cam.ac.uk

Matt Allen
*Department of Geography*
*University of Cambridge*
Cambridge, UK
mja78@cam.ac.uk

Carlos Cabo
*University of Oviedo*
Campus de Mieres, Spain
*Swansea University*
Swansea, UK
carloscabo@uniovi.es

Kim Calders
*Computational & Applied Vegetation Ecology, Department of Environment*
*Ghent University*
Ghent, Belgium
kim.calders@ugent.be

Amandine Debus
*Department of Geography*
*University of Cambridge*
Cambridge, UK
aed58@cam.ac.uk

Stuart W. D. Grieve
*School of Geography & Digital Environment Research Institute*
*Queen Mary University of London*
London, UK
s.grieve@qmul.ac.uk

Milto Miltiadou
*Department of Geography*
*University of Cambridge*
Cambridge, UK
mm2705@cam.ac.uk

Adam Noach
*Department of Geography*
*University of Cambridge*
Cambridge, UK
amn55@cam.ac.uk

Harry J. F. Owen
*Department of Geography*
*University of Cambridge*
Cambridge, UK
ho304@cam.ac.uk

Stefano Puliti
*Division of Forest and Forest Resources, Norwegian Institute for Bioeconomy Research*
Ås, Norway
stefano.puliti@nibio.no

*Abstract*— With the rise in high resolution remote sensing technologies there has been an explosion in the amount of data available for forest monitoring, and an accompanying growth in artificial intelligence applications to automatically derive forest properties of interest from these datasets. Many studies use their own data at small spatio-temporal scales, and demonstrate an application of an existing or adapted data science method for a particular task. This approach often involves intensive and time-consuming data collection and processing, but generates results restricted to specific ecosystems and sensor types. There is a lack of widespread acknowledgement of how the types and structures of data used affects performance and accuracy of analysis algorithms. To accelerate progress in the field more efficiently, benchmarking datasets upon which methods can be tested and compared are sorely needed.

Here, we discuss how lack of standardisation impacts confidence in estimation of key forest properties, and how considerations of data collection need to be accounted for in assessing method performance. We present pragmatic requirements and considerations for the creation of rigorous, useful benchmarking datasets for forest monitoring applications, and discuss how tools from modern data science can improve use of existing data. We list a set of example large-scale datasets that could contribute to benchmarking, and present a vision for how community-driven, representative benchmarking initiatives could benefit the field.

*Keywords—remote sensing, benchmarking, forests, artificial intelligence*

E. R. L., S. W. D. G, M. M. and H. J. F. O. were funded by a UKRI Future Leaders Fellowship (MR/T019832/1). E.R.L. was also supported by the Alan Turing Institute. M. A. was supported by the UKRI Centre for Doctoral Training in Application of Artificial Intelligence to the study of Environmental Risks (EP/S022961/1). C. C. was funded by a María Zambrano Grant (Spain; MU-21-UP2021_030) and the Natural Environmental Research Council (UK; NE/T001194/1) K.C was funded by the European Union (ERC-2021-STG Grant agreement No. 101039795). Views and opinions expressed are however those of the author(s) only and do not necessarily reflect those of the European Union or the European Research Council Executive Agency. Neither the European Union nor the granting authority can be held responsible for them. A. D. and A. N. were funded by the NERC C-CLEAR doctoral training programme (grant no. NE/S007164/1). S. P. was funded by the Center for Research-based Innovation SmartForest: Bringing Industry 4.0 to the Norwegian forest sector (NFR SFI project no. 309671, smartforest.no).

## I. Introduction

With increasing interest in, and monetisation of, the value of forest ecosystems for mitigation and adaptation to climate change, new pressures are growing for large-scale rapid, accurate and robust monitoring approaches [1]. Forests are home to up to 80% of terrestrial biodiversity [2], are a globally important store and sink of carbon [3], [4], and are therefore target ecosystems for enhancing carbon storage and sequestration, for biodiversity loss prevention, and for climate regulation. Accordingly, afforestation and forest restoration are central to many international initiatives (e.g. the Bonn Challenge [5], AFR100 [6] and ECCA30 [7]) and individual countries' National Action Plans [8].

Charting progress towards these goals, as well as effective monitoring of deforestation, degradation, and forest responses to climate change, relies on efficient large-scale, low-cost monitoring to, where possible, automate data collection and processing. In the last decade or so there has been a rapid increase in the availability and uptake of remote sensing technologies (from the ground, air and space) that are capable of collecting data at high enough resolution to identify individual trees. These new data sources have the potential to substantially increase the spatiotemporal scale of information available about the state and functioning of forests, compared to manual ground measurements, but come with challenges of interpretation [9]. Nevertheless, use of these sensors has led to an explosion in the size, type and complexity of datasets available, and therefore the need for new methods of analysis drawn from data science to rapidly extract critical, ecologically relevant, information from them. Further, the way in which the field has developed has brought particular challenges which make intercomparison of studies, and therefore methodological development, difficult. One particular reason for this is the common approach of using small-scale data for development due to a lack of widely accepted and representative benchmarking datasets. Here we present a perspective of what is to be gained by the creation and use of such datasets, and a vision for what properties they need to be of most use to the community.

While artificial intelligence (AI) shows impressive promise in tackling some of the big challenges of rapid, ecologically relevant, forest monitoring, including individual tree and species identification [10], biomass and carbon estimation [11], forest health and invasive species detection [12], and disturbance and degradation characterisation [13], method development has largely occurred in ad-hoc ways on small datasets. Data collection approaches in the discipline are not standardised, with differences in sensor choice and collection strategy often according to research group experience, equipment cost and availability, and ecosystem type. Differences in data resolution and quality are inherent in the ways in which they are collected in the field. Differences in data collected for the same task arise from: different types of sensors and differences within single sensor types; different ecosystems with highly varied levels and types of complexity (including structural, spectral, and diversity), and different approaches to collection and survey methods. Such heterogeneity in the structure of data means we can expect different AI methods to perform the same task very differently on different datasets [14], making easy choice of an approach impossible for a user. This confusion slows the development of robust and standardised approaches needed for the large-scale application of these technologies to solve pressing environmental problems.

Rapid increases in both data volume and processing capacity mean that the field can and should now switch from small, single-sensor, single-ecosystem method development to multi-sensor multi-ecosystem benchmark data sets upon which algorithms should be tested. The gains to be made from such a switch - through development of robust methods, reduced researcher effort and barriers to entry, and increased transferability of methods to new ecosystems and sensors - are substantial.

## II. Key applications of artificial intelligence for forest monitoring

High resolution remote sensing data analysed with AI methods have the potential to increase the information available for a wide range of forest monitoring applications by orders of magnitude [9], including to automate at large scale knowledge about properties currently understood only at small spatial scale with manual approaches. Here, we identify a series of use cases where the impact of widespread uptake of use of artificial intelligence methods with such data will be high, and where methodological development is commonly carried out on small sample sizes, making assessing the value of a method for a different sensor or ecosystem difficult, and so for which use of benchmarking dataset is particularly important:

- *Individual tree detection and extraction:* the monitoring of individual trees is key to understanding forest diversity, demographics, and dynamics that inform long-term resilience. The extraction of individuals remains a major bottleneck to large-scale monitoring using high resolution remote sensing, and even the most widely adopted rules-based approaches require manual refinement [15]. Because of their importance for so many applications of these data, the problems of individual tree detection and extraction have attracted a significant amount of interest for the application of deep learning solutions, including identifying individuals using bounding boxes on tree crowns in 2D imagery (e.g. [10], [16]), and tree crown extraction from 3D airborne LiDAR (ALS) (e.g. [17]).

- *Point cloud semantic segmentation:* The separation of three-dimensional point clouds into leaves, wood, deadwood, ground and other material classes is a crucial step towards understanding ecosystem structure and function. For example, wood-only point clouds can provide representative extrapolations of point based wood respiration measurements [18], accurate estimates of total wood volume [19], and tree hydraulic structure [20]. Accurate estimates of leaf density, angles and arrangement from point cloud data [21] allow quantification of light microenvironments within crowns and across canopies [22]. Two broad approaches to semantic segmentation have emerged: either using predefined features fed into machine learning models (e.g. [23], [24]), or deep learning classification of raw point clouds (e.g. [25]).

- *Species detection/classification:* individual species identification is crucial to monitor biodiversity, to estimate carbon storage, and to understand forest dynamics that depend on interspecific interactions, such as competition. Species identification currently relies on labour-intensive stem maps generated by field surveys, so AI-based solutions to derive these labels have attracted some attention, including methods based on terrestrial laser scanning (TLS; [14], [26]), ALS [27], high resolution 2D imagery [28], [29], and co-located aerial and satellite imagery [30].

- *Forest health and invasive species:* global change is increasing the rate, intensity and number of disturbances affecting forests around the world, including invasive pests and pathogens, and abiotic stressors such as increasing drought intensity [31]. Disturbances can cause large scale canopy tree dieback, a cascade of negative impacts on whole-ecosystem functioning and diversity [32], and a rapid loss of forest carbon [33]. Airborne sensors, mounted on unmanned aerial vehicles (UAVs/drones) or manned aircraft, offer the opportunity to rapidly survey large areas to identify trees suffering dieback to provide both early warning and inform protective management practices. Dieback can often be earliest identified by leaf stress, which creates small changes in leaf spectral properties and may be identified using 2D multi- or hyper- spectral imagery [34].

- *Carbon stocks:* rapid, large-scale robust estimates of forest biomass (and carbon) are key to carbon financing, but are notoriously difficult to accurately quantify. High resolution remote sensing, particularly LiDAR, has been show to produce very precise estimates of forest volume [19], and UAV imagery has been used to quantify agroforestry carbon using locally trained AI [1]. Recent efforts have demonstrated the availability of globally representative high-quality ground-monitoring sites, where benchmarking datasets might be developed [35].

- *Anthropogenic disturbance and degradation characterisation:* anthropogenic activities such as selective logging, fuelwood extraction, shifting cultivation, and over-hunting are major sources of carbon emissions [36] and impact forest structure, diversity, functioning, and carbon storage [37]. Identifying degradation and determining the nature of disturbances is crucial to understanding their effects on forests. Optical and SAR satellite data offer opportunities to detect and characterise forest disturbances using machine learning methods [38], for example, to identify selective logging (e.g. [13]). However, detection of smaller-scale events, especially with limited impacts on canopy cover or aboveground biomass, and finer-scale impacts on forest properties likely needs sensors with higher spatial resolution and/or canopy penetration capabilities [39], such as ALS (used in [40] to monitor selective logging), TLS (used in [41] to study the effects of degradation on forest structure) and imagery from the ground [42] or UAVs (used in [43] to study the effects of degradation on canopy texture).

III. BENCHMARKING IN FOREST MONITORING NEEDS DIVERSE, FUTURE-PROOFED DATASETS

Development of new AI approaches for the kinds of use cases discussed above should explicitly recognise the limited weight of single-sensor and single-ecosystem studies, and move to testing on data types representative of the diversity of both forests and sensors. Editors and reviewers have a particular role to play here, to ensure careful and caveated wording of studies' generalisability, to push authors to broaden the data they work with, and to encourage, support and promote the creation of appropriate benchmarking datasets emerging from the community.

Sensor development is rapid, and the nature of forest monitoring as a discipline means that, whilst some campaigns are large-scale and standardised (e.g. national ALS, [44]), large amounts of valuable data are collected at small scale. To require standardised data collection for inclusion in benchmarking would not only significantly reduce the data available, but also increase risk of future redundancies as sensor resolution increases. Instead, we propose that benchmarking datasets should include data from multiple sensors, including from different platforms, to promote the development of methods that are sensor agnostic and therefore as widely applicable as possible (e.g. [25]).

Methods should be developed and tested on existing datasets to calibrate algorithmic performance, and careful conclusions drawn that acknowledge limitations to implementation on the basis of untested data and forest types. Forest ecosystems vary enormously in structural complexity, so the same data collection techniques can result in data with highly variable amounts of information depending on the location and ecosystem type. For three-dimensional data from sensors like LiDAR, this is most markedly because of greater occlusion in complex canopies [9], leading to much less information about structural elements further from the sensor (for ground based, lower density canopy data, and for airborne, lower density ground and trunk data), whereas for two-dimensional imagery, multi-layered forests with high stem density may have less clear discrimination between crowns, making crown and tree detection much harder. There is a mismatch between the geographical location of the highest diversity and highest carbon forests, and dataset coverage for key forest properties [35], and ground truth data for many use cases relies on diverse and highly specialised data collection that is not always well recognised or rewarded (see [45] for a discussion of fair rewards for grassroots communities collecting tropical forest data). Benchmarking datasets must fully acknowledge all contributors, and where possible connect with initiatives collecting and processing ground truth information.

Although high-quality ground truth labels are clearly necessary for performance comparison, unlabelled data is not without value. In particular, for computer vision related tasks, self-supervised pre-training methods - including contrastive [46], reconstruction [47] and self-distillation-based ([48], [49]) methods - have achieved substantial success in transfer learning without the need for ground truth labels at the pre-training stage. Such pre-training methods enable leveraging of high-quality unlabelled data - which is often obtained in huge volumes from fieldwork campaigns - to achieve performance approaching the fully supervised case, while using only a fraction of the labels. These methods are of particular value in forest monitoring, where data labelling, particularly for point-based three dimensional data, is extremely time consuming. Reducing reliance on the burdensome labelling process from the data processing pipeline would allow improved forest monitoring at larger scale, and in a wider variety of hard-to-reach locations. Particularly promising are the possibilities these methods bring to enable the use of smaller, site-specific datasets coupled with deep learning models for which large data requirements would be otherwise restrictive.

IV. BENCHMARKING IN FOREST MONITORING NEEDS COMMON EVALUATION METRICS

Adopting common benchmarking datasets is not enough to monitor methodological development - we also need to adopt common evaluation metrics. In particular, the use of common performance metrics is crucial when evaluating whether algorithm performance transfers from standard AI benchmark datasets to ecological data. For example, the performance of individual tree crown (ITC) detection algorithms are commonly reported in ecological literature using precision, recall or F-score [50], [51], but the standard metric used to evaluate performance on the ubiquitous COCO benchmark [52] is mean average precision (mAP). Mean average precision is likely to select models that are more robust across confidence and Intersection over Union (IoU) thresholds, and should be adopted to measure performance on ecological data - although there is scope for discussion on whether the exact COCO definitions of mAP are appropriate for model selection in this use case (for example, whether averaging from IoU thresholds as low as 0.5 is desirable for ITC detection in dense canopy). Regardless, the calculation of such metrics is still necessary for cross-discipline comparison.

To enable such consistency it is important to develop standard benchmarking systems that allow for fully independent evaluation based on a set of defined metrics. An example of such an effort can be found in the NEON tree evaluation benchmark [53], where a user can independently evaluate tree bounding-box detectors based on a predefined set of metrics on multi-sensor data (co-registered RGB, LiDAR and hyperspectral imagery).

## V. CONSIDERATIONS FOR SELECTING FIT FOR PURPOSE DATASETS

Selecting a dataset for benchmarking purposes for a given task requires consideration of the data collection methods, an understanding of the structure of the data, and clarity on the scope of the task. Depending on the sensor used, relevant considerations for assessing the suitability of a high resolution remote sensing dataset for a forest monitoring task will depend on: the locations and ecosystem types covered; the conditions of the sites and the time of year; the sensor used; the surveying approach (e.g. flight height, line density or ground scanning density, and survey team expertise); the data processing, including geographic referencing and precision, orthorectification methods applied, and any potential data distortions introduced, for example through image alignment.

## VI. REQUIRED AND DESIRABLE PROPERTIES OF BENCHMARKING DATASETS

Following the considerations of the previous section, and in line with recommendations in related domains [54], we suggest minimum required and desirable properties of high resolution remote sensing datasets to be considered suitable for benchmarking in the types of use-cases of forest monitoring discussed here.

*Required*
1) Open with unrestricted access.
2) Sufficient metadata including sensor types, point resolution, date of collection, ecosystem description (including managed/unmanaged, leaf-on/off), collection strategy, geographic reference system, precision of location information, with comprehensive contributor acknowledgement.
3) Clearly defined tasks for which data are suitable, with relevant ground truth information.
4) Defined evaluation metrics.

*Desirable*
5) Demonstration notebooks available for data access and visualisation, procedures for data down-sampling (point density or image resolution), and example artificial intelligence implementations.
6) Multi-ecosystem data from multiple, co-located sensors, to facilitate multi-model model development (e.g. point cloud with image/hyperspectral data).

## VII. BEST PRACTICE FOR THE USE OF BENCHMARKING DATASETS

A goal of the field must be to collate and standardise the use of high quality, well documented data that, for a given use case, is representative of as wide a variety of types and conditions of ecosystems as possible, and which is accompanied by high-quality ground truth information to validate accuracy of methods and quantify uncertainty of outputs. However, just as important is clear information on use and limits of the data, standardised file types and formats, and standard processing pipelines, such that the barriers to participation are reduced, and high impact interdisciplinary development and collaboration is encouraged.

Suitable use cases should be justified according to tests of the quality of the data, and inherent limitations of application should be clear. However, data manipulation and simulation approaches can enhance existing datasets to create better benchmarking. For example, multi-sensor data will improve future-proofed, sensor agnostic, and collection method agnostic model development (e.g. [25]), however down-sampling methods applied to existing data can simulate different collection strategies and sensors. Model performance compared against multi-sensor data, real or simulated, can also feed back into informing further fieldwork, including cost-effective data collection for different use cases, and even direct future sensor development. Where choice of method depends on the sensor or structure of data, multi-sensor data can inform the creation of multi-pass classifiers, with the first pass classifying the type of data or forest, and the choice of model or classifier for the second task selected according to the best performing algorithm given the data.

Manual labelling using ground truth information, whilst crucial for model validation, is time consuming, and strict requirements will always limit data size. Instead, simulated data can massively increase data size available for training, and, as long as they are presented alongside guidance on representative forest types, are a valuable source of training information (e.g. [55]). Further, simulated sensor data for unobserved or unmeasured use-cases might be generated using radiative transfer approaches applied to dynamic digital twin models.

Just as performance evaluation metrics require standardisation, test/train/validation sets of benchmark datasets must also be established for effective model comparison. Variability in structural and spectral complexity in remote sensing datasets of forests means data splits for model evaluation should be representative across ecosystems within the dataset (ideally, globally). However, splits should not be permanently enforced in a way that would prevent the accumulation of new sensor or ecosystem data that may come on line after initial benchmark creation. In addition, while model developers may focus on improving model generality and transferability, model *users* are likely to favour the model that performs the best for their specific ecosystem, use case, and quality of data. One pragmatic way to acknowledge this is to ensure that performance evaluation is reported with as much detail as possible, for example, for a tree crown detection model, performance could be reported both across the whole dataset, and contextually, for each ecosystem type.

## VIII. EXISTING LABELLED DATASETS FOR BENCHMARKING

We searched for benchmarking datasets exhibiting the properties discussed above, with ground truth or manually labelled data, to give as examples for our use-cases, and give examples in Table I.

Although this list is by no means exhaustive, in surveying for benchmarking examples we noted many exciting datasets that fail in one or more of our required criteria, but which, with some targeted effort, would be suitable for this need. Notwithstanding acknowledgement of the excellent efforts listed above, we found few datasets representing more than one ecosystem type. Therefore, rather than suggesting these as sufficient for the listed use cases, we instead suggest they provide example approaches towards benchmark creation, and could form a basis of and model for wider efforts.

TABLE I.

| Dataset name | Sensor and ground data types column subhead | Ecosystem coverage | Use cases |
|---|---|---|---|
| ReforesTree [1] | UAV RGB images. Individual tree locations, crown bounding boxes, DBH, species, biomass. | Dry tropical agroforestry in Ecuador, >4400 trees, 3.17 hectares. 28 species. | Carbon stocks |
| TreeSatAI Benchmark Archive [30] | Co-located Sentinel 1 and 2, airborne RGB+NIR imagery. Species/age labelled patches (>50k 60x60m image triples: S1, S2, aerial). | Temperate forests in Lower Saxony, Germany. 20 species of 15 genera. | Species identification |
| Barknet 1.0 [29] | 23,000 high resolution bark images with ground species identification. | Parks and forests in Quebec, Canada. 23 different tree species. | Species identification |
| NeonTreeEvaluation Benchmark data [10] | Co-registered ALS, hyperspectral and RGB data. 3000 field stem locations, and >400 field-annotated crowns. | 22 NEON sites across mainland USA. | Tree detection |
| Urban Tree Detection Data [16] | Multi-spectral aerial imagery (60cm pixel). c. 100k image-annotated tree locations. | Urban areas across southern California. | Tree detection |
| DeepFire dataset [42] | RGB multi angle photographs of forests with standard resolution, manually classified fire/no fire. | Images of forests and mountainous regions obtained from online search. | Forest fire detection |

## IX. CONCLUSIONS

We found many more published examples of proposed benchmarking datasets in the literature than those listed in Table 1, but most cover only a handful of plots and so do not represent spatial scales large enough to capture structural and spectral diversity, represent only one forest and one sensor type, and/or may not include sufficient information on the reliability of ground truth information, or errors related to derived forest properties. Such small scale, restricted datasets are not suitable for the development of widely applicable AI approaches, but large-scale multi-sensor datasets are expensive to generate and rare, so waiting for such initiatives is not feasible.

Instead, we believe that existing data, carefully curated, should form the basis of community-created and collated benchmarking datasets. Sufficient meta-data to determine the value of datasets for particular use-cases is vital, and so proper acknowledgement of contributors is crucial to incentivise what can be a thankless additional task. For many ecosystems and use cases the collection of more data is less important than the collation and curation of what already exists, particularly when large-scale data may exist but is not yet curated for our benchmarking needs (e.g. [56]). We believe that whilst this requires some effort of data sharing and processing, the pay-off will be worth the community effort, and that such curated, diverse and geographically representative benchmarking data will significantly accelerate forest monitoring capabilities.

ACKNOWLEDGMENT

We thank the 3DForEcoTech COST Action CA20118 for facilitating discussions that contributed towards this paper.